# CoSeNet: A Novel Approach for Optimal Segmentation of Correlation Matrices

A. Palomo-Alonso[a], D. Casillas-Pérez[b], S. Jiménez-Fernández[a], A. Portilla-Figueras[a], S. Salcedo-Sanz[a]

[a]*Department of Signal Processing and Communications, Ctra. Madrid-Barcelona, km 33, Alcalá de Henares, 28805, Madrid, Spain*
[b]*Department of Signal Processing and Communications, Camino del Molino, 5, Fuenlabrada, 28942, Madrid, Spain*

**Abstract**

In this paper, we propose a novel approach for the optimal identification of correlated segments in noisy correlation matrices. The proposed model is known as CoSeNet (Correlation Segmentation Network) and is based on a four-layer algorithmic architecture that includes several processing layers: input, formatting, re-scaling, and segmentation layer. The proposed model can e!ectively identify correlated segments in such matrices, better than previous approaches for similar problems. Internally, the proposed model utilizes an overlapping technique and uses pre-trained Machine Learning (ML) algorithms, which makes it robust and generalizable. CoSeNet approach also includes a method that optimizes the parameters of the re-scaling layer using a heuristic algorithm and fitness based on a Window Di!erence-based metric. The output of the model is a binary noise-free matrix representing optimal segmentation as well as its segmentation points and can be used in a variety of applications, obtaining compromise solutions between e"ciency, memory, and speed of the proposed deployment model.

## 1. Introduction

In today's world, an immense amount of data is processed every day. In many applications, it is crucial to identify relationships between di!erent elements and group them accordingly [1]. In many cases, these relationships are indicated by a correlation function [2, 3], which measures how correlated two elements are, based on one or more characteristics or metrics. By applying a correlation function to every combination of elements, it is possible to generate a square correlation matrix with a length equal to the number of elements being processed. However, noisy correlation matrices may be produced if the correlation function is not e"cient enough, or if the elements do not contain enough information about the characteristic that links them. Thus, the problem of segmentation of correlation matrices consists of, giving a series of elements in a correlation matrix, belonging to one or more groups and spatially ordered, and given a function capable of obtaining a value (metric) from the relationship among these elements, we seek to obtain (or identify) the points that correctly separate these groups.

The detection of correlated segments in noisy correlation matrices is an important problem, closely related to the well-known subspace clustering problem [4, 5], that appears in di!erent fields, mainly associated with the discovery of underlying patterns and relationships in complex data sets. This problem has been massively treated in image processing [6, 7], in Natural

Language Processing [8], and in many other research fields. In Finance, for instance, correlation matrix segmentation is used to analyze clusters for financial data [9], which can be useful for portfolio diversification, risk management, and asset correlation [10, 11]. In Biology, the segmentation of correlation matrices has been used to identify co-regulated genes and infer cis-regulatory modules [12], which can provide insight into gene function and regulation [13], and also in biomedical applications with a signal processing component, such as human vessel segmentation [6] or heart sounds analysis [14]. Moreover, in Physics, correlation matrix segmentation has been used to identify communities or groups of interacting particles in complex systems, such as social networks and biological networks [15]. This can provide insight into the underlying structure and dynamics of these systems. In Climate science, correlation matrix segmentation has been used to identify patterns in climate time series data, such as El Niño-Southern Oscillation (ENSO) and the North Atlantic Oscillation (NAO) [16], which can be used to improve climate predictions and understand the dynamics of the Earth's climate system.

Previous works have addressed this problem using di!erent classical techniques such as clustering algorithms [17, 18, 19], statistical models [20, 21] or graph-based algorithms [22]. However, many of these methods have limitations when dealing with highly noisy and imperfect correlation functions. More recently, several studies have introduced specific variations of the classical algorithms, which work better in these cases of noisy correlation, for example, *Correlation Clustering* [23], *Community Detection* (CD) techniques [24] (based on graph algorithms and graph analysis techniques), and *Deep Clustering* techniques [25]. More recent approaches revisited Hierarchical Clustering approaches [26] and Modularity maximization (MM) [27], with good results in the detection of correlated segments in noisy correlation matrices.

In this study, we propose a novel approach for the optimal segmentation of correlation matrices, based on a complete sequential multi-algorithm architecture that involves di!erent processing levels, each implementing several methods. Specifically, the proposed approach consists of three layers, which can be grouped into three categories. The first level consists of di!erent procedures essential to optimally prepare the input data. The second level, *Metaheuristic*, adapts the input and output data using classical algorithms. Finally, the third level is formed by di!erent Machine Learning (ML) algorithms able to accurately identify the boundaries in the provided correlation matrix. Thus, the proposed multi-algorithm architecture can process square correlation matrices of any scale and size, using a ML model capable of identifying correlated segments with high performance, even for highly noisy data. The proposed approach can adapt any matrix, regardless of its size, to the ML model with excellent performance. The proposed approach also runs faster on general-purpose processors, making it a more practical solution for real-world applications. Note that, unlike the rest of the algorithms in the state of the art, the proposed architecture implements simple ML algorithms together with heuristic and other methods. The main di!erence with the previous approach is that our method can accurately solve the problem with these simple ML algorithms, so the error rate, execution speed, and memory size improve with respect to alternative (heavier) algorithms. Additionally, the architecture employs a heuristic to perform custom fine-tuning of the algorithm's parameters.

The algorithm's performance has been assessed over a highly nonlinear and noisy database. The problem proposed in the comparative is a problem of text segmentation by topics. We obtain random articles from Wikipedia, concatenate them, and divide them into sentences. With a Language Model (BERT [28]) we generate a sentence similarity coe"cient, used as correlation value, and correlation matrices are generated with these values sentence by sentence. The e!ectiveness in identifying correlated group segmentation and its superiority to some state-of-the-art algorithms such as unsupervised, Community Detection, and Deep Clustering have been tested, reaching improvements of 6% - 22% in terms of performance. The proposed approach aims to propose a unified solution to the problem, with the possibility of performing fine-tuning with a few samples from the database. We have also included a GitHub repository containing the source code and all the experiments in this article, as well as a PyPi package for Python versions higher than 3.8 that can automatically perform matrix segmentation using the proposed model.

The remainder of the paper has been structured as follows: the next section presents the proposed approach for correlation matrices segmentation, following a sequential order according to its different layers. Within the approach, different ML techniques have been implemented as final prediction models. Section 3, discusses what is the best ML model for prediction following a comparison of several pre-trained ML models to a synthetic database. In this section, we also provide details on the databases used and a brief explanation of the candidate ML algorithms. Section 4 shows the results obtained by the proposed optimized model for a real problem database, comparing the results obtained with those by different state-of-the-art algorithms addressing the same problem. Finally, Section 5 closes the paper with some conclusions and remarks on the research carried out.

## 2. CoSeNet: proposed multi-algorithm architecture

This work aims to present a generalized architecture (CoSeNet) able to detect groups of correlated information in matrices (therefore performing a matrix segmentation), with application in a broad class of different problems, including noisy situations where grouping is difficult to obtain. Thus, the model's input is the information regarding the relationship among the elements in a given problem, so a correlation matrix (or any information that can be transformed into that) will be used. The model's output is, of course, the segmentation performed.

The proposed CoSeNet approach consists of three processing levels, each formed by an input and an output layer, except for the last level, which only presents one layer (see Figure 1).

Level 1 provides generality to the model in terms of the input's matrix size. That is, the CoSeNet model needs to segment correlation matrices independently of their size, as algorithms used for segmentation cannot always work with variable sizes. Therefore, Level 1, Layer 1 (input layer) receives a noisy correlation matrix of a given size and splits and pads it into sub-matrices, if necessary, preparing it for further processing. Therefore, it provides generality to the model in terms of the input's matrix size. Then, Level 2, Layer 2 provides generality to the model in terms of the input's scale (range values). That is, the CoSeNet model needs to be independent of the use case, and each correlation problem may be provided on a different scale. For this purpose, *metaheuristic* approaches are used both in the input and output layers to solve this challenge. Finally, Level 3 performs the segmentation. As there are many different *Machine Learning* approaches in the literature for this purpose, the model can operate with many of these: CoSeNet offers generality by providing a pre-trained model based on the use of a synthetic database, that incorporates different noisy situations, used to identify the boundaries of each group, yielding to the final segmentation. Therefore, the user does not have to pre-train the chosen model for the specific problem but can tune parameters to adapt the model to the specific problem requirements. The output layers are devoted to providing a noise-free segmented matrix and its corresponding segmentation vector (that indicates the predicted segmentation points) both matching the original input matrix. Subsection 2.1 explains the input correlation matrix specifications as well as the output segmentation vector. Subsections 2.2 to 2.4 explain in detail all the levels of the CoSeNet model.

### 2.1. Input data, problem encoding and output data

CoSeNet approach receives as input a normalized correlation matrix (see the left side of Figure 1). We assume that the correlation function is symmetric and square, meaning that the correlation of one element with another is reciprocal and yields the same value if the positions of the elements in the arguments of the function are interchanged.

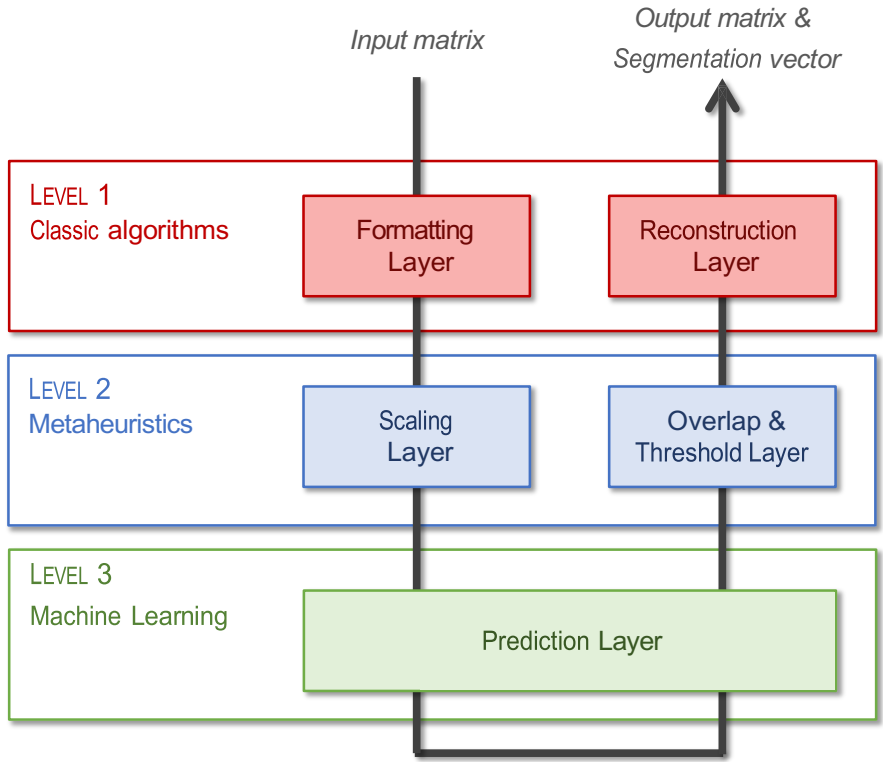

Figure 1: Proposed multi-algorithm architecture with the description of each level, for optimal noisy matrices segmentation.

Three key parameters define this matrix. First, the size of the input matrix is not previously set (CoSeNet generalizes all possible input problems), but it has to be square. Second, the scale of values in the correlation matrices varies depending on the correlation function used for the problem at hand. For instance, a high ratio between two elements for one correlation function problem may be a value bigger than 0.3 (in the range [0.0-1.0]), while another correlation function may identify a high correlation with a value of 0.9 and low correlation with 0.3 (using the same range [0.0 - 1.0]). Thus, having to generalize the use of di!erent correlation functions is a challenge that has to be solved. Finally, the number of correlated segments that need to be obtained from the matrices is unknown and may vary from one problem to another. However, in the problem's context, these matrices must have an implicit spatial correlation. That is, they are spatially ordered.

The input matrix will be referred to as $\mathbf{R}_{in}$, and the original size of the input matrix is denoted as $M_{in}$. Also, we will refer to the scale of the correlation function as scale $A = scl(\mathbf{R}_{in})$, where the function $scl$ refers to the input's scale value.

In the context of the problem, we need to adapt the input matrices to a set size because the segmentation algorithms used at the architecture's last level may only process arrays of a fixed size. For instance, the majority of ML models cannot work with variable input sizes: a Neural Network cannot vary the size of input neurons, or a linear regressor cannot vary the number of input dimensions. However, our system must globally be able to process matrices of di!erent sizes. Therefore, we refer to the segmentation algorithms' set size as the "throughput" ($T$) of the system, which is a crucial parameter for the whole approach (without loss of generality, in this work we have considered $T$ even).

Figure 2 illustrates an example of a correlation matrix (normalized in the range [0, 1]) and its segmentation into a previously unknown number of groups. The first eight rows by eight columns represent the correlation matrix itself (red color represents a high correlation value and black color represents a low correlation value), while the bottom row presents the segmentation (the green cell corresponds to the beginning of each segment). Figure 2a presents an example of a correlation matrix that contains two distinct groups of highly correlated elements, where the red color represents the value of the correlation function in the matrix, the brighter it is, the larger the relationship between the elements. The first group consists of elements 0 to 3, and the second group consists of elements 4 to 7. In this case, the correlation function is well-defined, and there is a small amount of noise in the matrix. However, Figure 2b shows an example where the correlation between elements is not as clear. In this case, there are three groups (elements 1 to 3, 4 to 5, and 6 to 8), and there is noise in the matrix (the boundaries between two different groups are not clear).

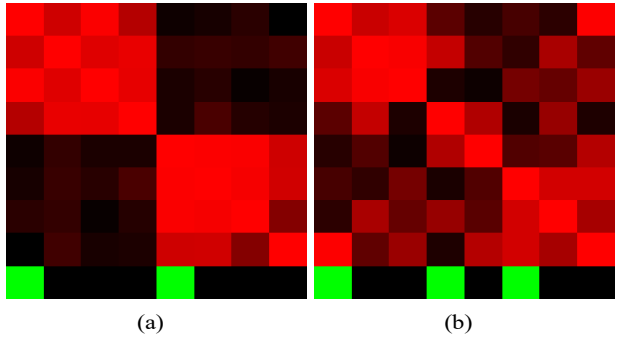

Figure 2: Example of correlation matrices (in black and red colors) and their associated segmentation (in black and green, with a green cell at the starting of each segment); (a) Almost noise-free matrix with two strongly correlated segments; (b) Noisy matrix with three correlated segments.

In this work, for a correlation matrix of size $M_{in} \times M_{in}$, the output segmentation vector ($\mathbf{s}_{out}$) is a binary vector of size $1 \times M_{in}$.

In this output binary vector, a value of 0 indicates that the corresponding index does not correspond to the start of a new group of elements, while a value of 1 indicates that a new group of elements begins at that index. Since the number of groups is previously unknown, it is important to highlight that this number is determined by the segmentation algorithms. Therefore, this is a regression problem for the third-level algorithms, since the algorithms must estimate the segmentation vector which will be denoted as $\mathbf{s_i}$.

*2.2. Level 1, layer 1: Formatting the input data*

The Formatting Layer is a Level 1 layer (See Figure 1) located at the input (see the left side of Figure 3). Its principal purpose is to split the input matrix into a given number of sub-matrices, to provide generality to the proposed model. This is needed due to two facts: 1) different applications are defined by different sized inputs, and 2) different Machine Learning solutions (explained in Level 3, Subsection 2.4) can be applied to solve the segmentation prediction, and each one of them may need a specific or fixed size. Therefore, the Formatting Layer receives an input matrix of size $M_{in}$ and divides it into $V$ sub-matrices of size $T$. Note that $V$ depends on the matrix's input size and the throughput $T$.

This splitting process is achieved with a technique we denote as "Window Overlapping Copy on the Diagonal (*WOCD*)". To perform this technique, it is necessary that the original size of the input matrix $M_{in}$ is divisible by $T/2$. Since this only happens in exceptional cases, a process to expand the input matrix to size $M_0$ ($M_0 \geq M_{in}$) is needed. The procedure used to expand this matrix is denoted as "Identity Padding" (*IP*), and consists of extending the original input matrix until its size $M_0$ is divisible by $T/2$ and filling it with zeros, except for the main diagonal, were it is filled with 1 (as if it was the identity matrix). Once the padded matrix has been expanded, we split it using the *WOCD* technique.

The *WOCD* method involves copying sub-matrices of size $T \times T$ every $T/2$ elements onto the main diagonal $V$ times (note that $V$ is the number of sub-matrices present). Although it would be possible to copy the sub-matrices every $T$ elements, it is necessary to overlap them to ensure that the segmentation at the boundaries of each matrix is optimally reconstructed (that is, to provide context information). Figure 4 illustrates this overlap, which is critical, as the points where the segmentation is more likely to fail are those corresponding to the edges of the sub-matrices, while the easiest points to predict are in the center.

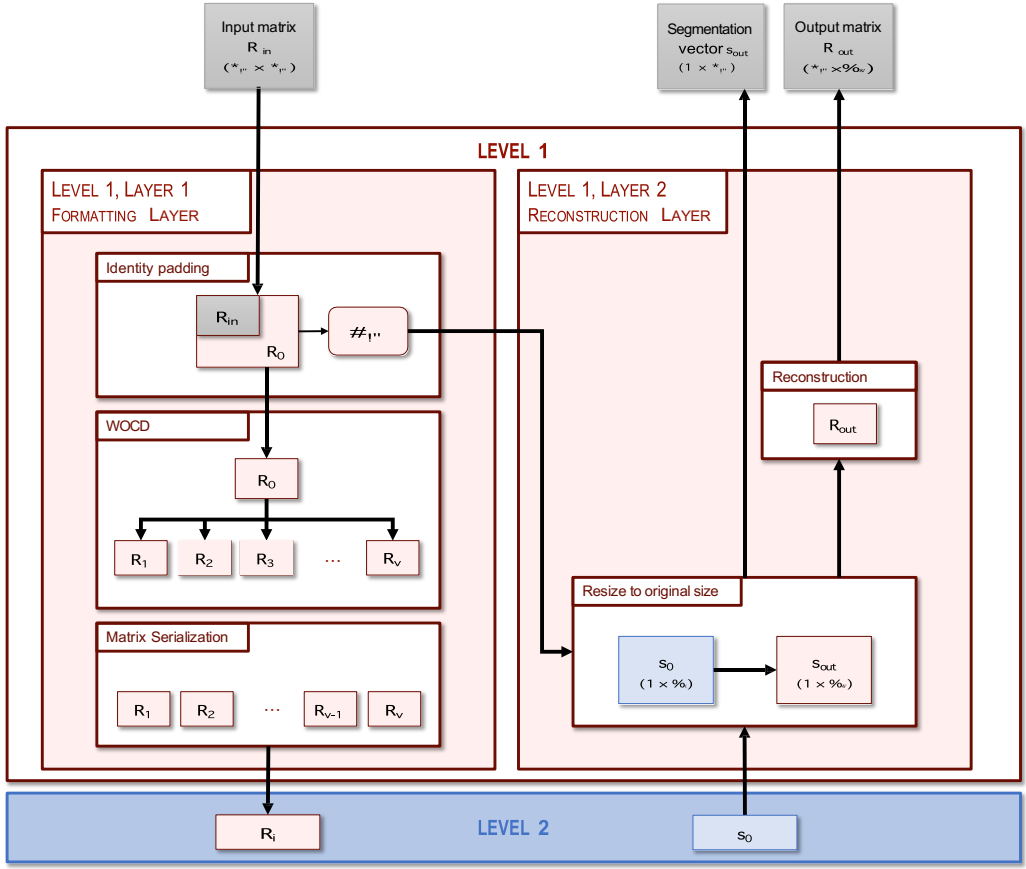

Figure 3: Classical processes involved in the model's first level (Level 1). Formatting Layer (left side, Layer 1) and Reconstruction Layer (right side, Layer 2).

It is important to consider the relationship between the model's throughput and the input matrices and the estimated size of the element groups. If the number of elements in a group is larger than the throughput, it is still acceptable and the proposed architecture will be able to perform the segmentation. Nevertheless, if the throughput size is too unbalanced, the system's performance may decrease slightly.

The number of matrices and the outgoing length of the input matrix resulting from applying $WOCD$ on the input matrix with $IP$ is given by Equation (1):

$$V = \begin{cases} \left\lceil \frac{2 \cdot M_{in}}{T} \right\rceil - 1, & \text{for } M_{in} \geq T \\ 1, & \text{for } M_{in} < T \end{cases}$$

$$M_0 = \frac{T \cdot (V + 1)}{2}$$

(1)

Once the $WOCD$ is performed on the expanded matrix, the resulting matrices are serialized, ordered, and handed to the next layer. The original size $M_{in}$ of the input matrix is stored for later reconstruction.

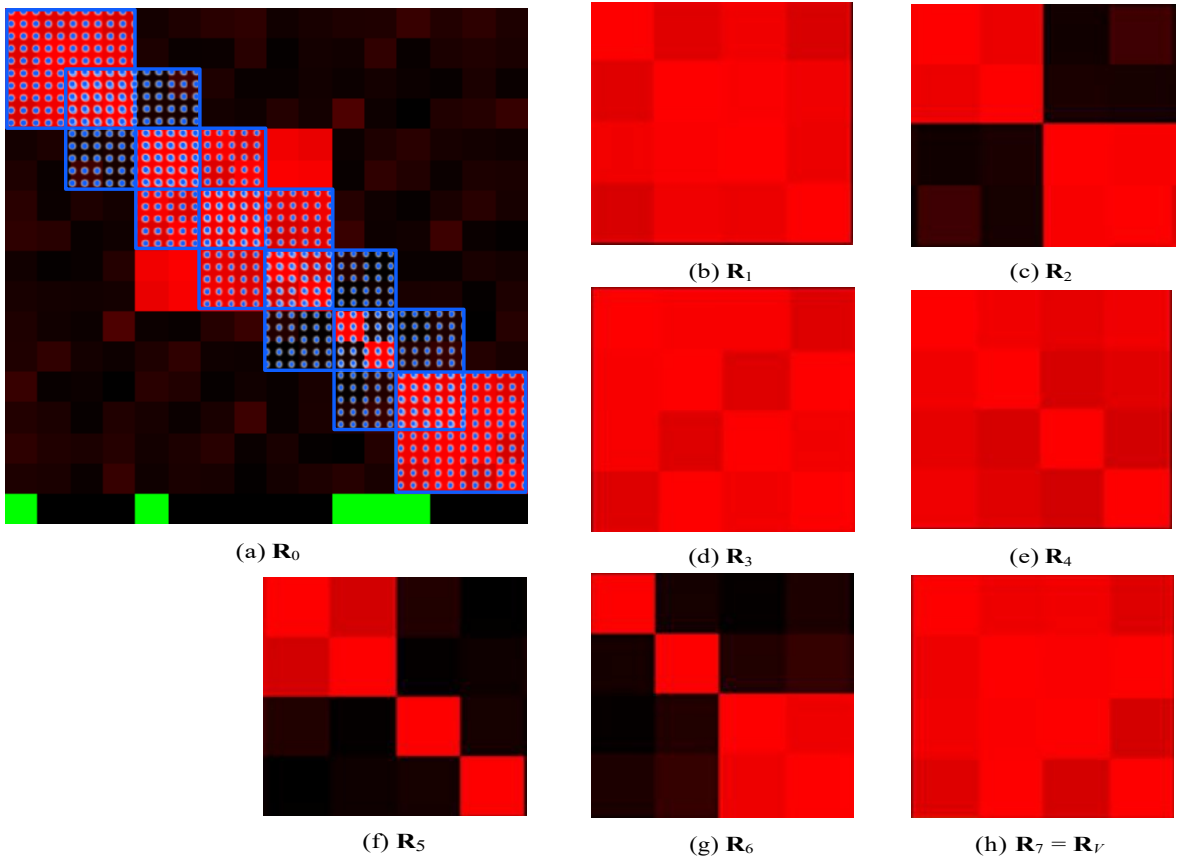

Figure 4: Window Overlapping Copy on the Diagonal (*WOCD*) process for a 16 → 16 matrix with throughput $T = 4$ and the corresponding sub-matrices generated: $R_1$ to $R_7$; where (a) shows the expanded matrix ($R_0$ in red), (b) to (h) are the copies to overlap (shown also superimposed in (a) in blue).

*2.3. Level 2, Layer 1: Scaling layer*

The scaling layer is a Level 2 layer (see Figure 1), and its main purpose is to prepare di!erent problems' matrices to fit into a unique scale, to provide generality to the proposed CoSeNet model. Note that this process can be indistinctly addressed before or after dividing the initial matrix into sub-matrices, that is Level 1, Layer 1, and Level 2, Layer 1 can be interchanged. Without loss of generality, in this work, we propose fixing the scaling problem at the sub-matrix level of the system (Level 2, Layer 1).

Figure 5 shows the model's second level. On the left side, a schematic representation of Level 2, Layer 1 (responsible for addressing the scaling problem) is presented.

The scaling problem is needed as the ML model used in Level 3 is designed on a given scale (referred to as *scale B*) between two values (typically 0-1). When values of the correlation matrix used in the specific problem are provided on a di!erent scale (referred to as *scale A*), segmentation errors may occur, since the ML model has been built to consider *scale B*. The scaling problem is an important challenge associated with the correlation function used to generate the matrix and is highly dependent on the specific problem at hand.

Thus, to generalize the model, a re-scaling function composed of several terms has been considered in this work, as di!erent re-scaling functions may be optimum for di!erent problems. In all cases, parameters determining each one's weight for the specific problem at hand have to be discovered.

Let us define a re-scaling function $f_{scale}(\mathbf{R}, \boldsymbol{\omega})$ defined to transform the values of the input matrix $\mathbf{R}$ to a new *scale $\hat{A}$*. The re-scaling function parameters are represented by $\boldsymbol{\omega}$.

To determine the parameters that best fit the problem, we implement an evolutionary algorithm. Thus, we reserve a small amount of the data as validation data and it is used only to optimize these parameters. Take into account that we seek to minimize the difference in scales $\arg\min_{\omega}\{\hat{A} - B\}$ or maximize the system's performance $\arg\min_{\omega}\{MSE(\hat{s}_{out}, s_{out})\}$ (where $MSE$ is the Mean Squared Error) given a prediction model in Level 3.

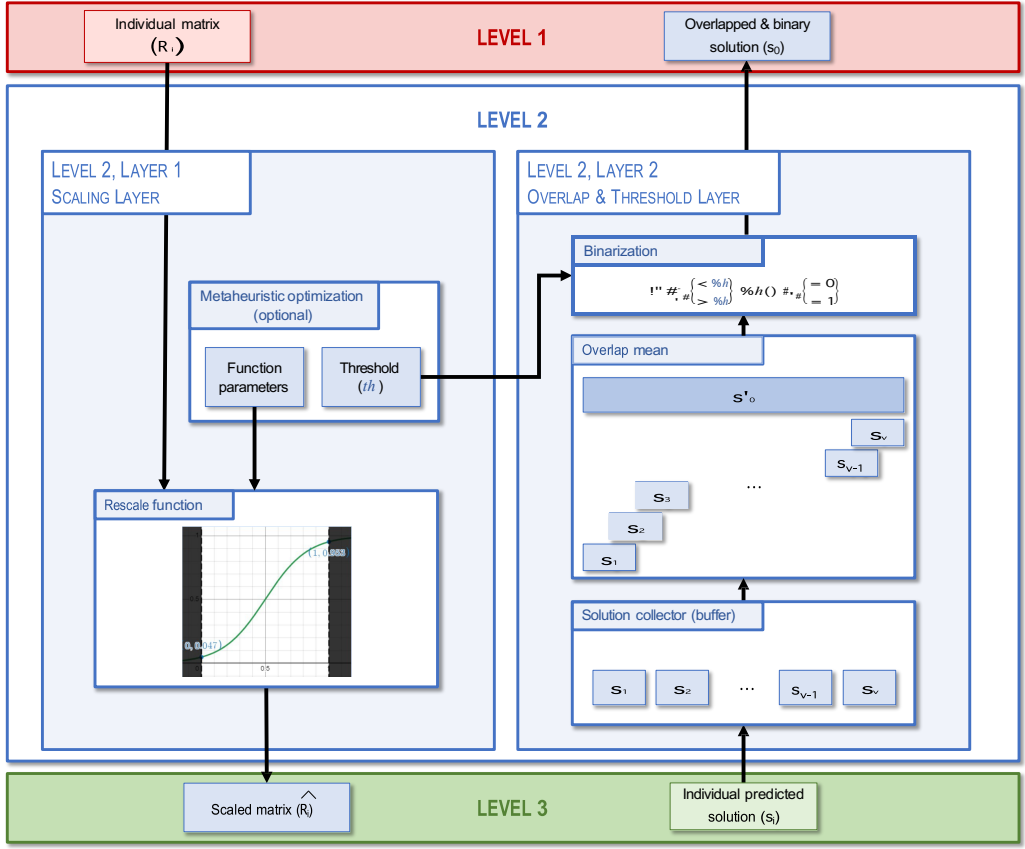

Figure 5: Heuristic processes involved in the model's Level 2. The left side presents Layer 1 (Scaling Layer) while the right side presents Layer 2 (Overlap & Threshold Layer).

In this work, we suggest the use of Equation (2) as a re-scaling function to produce $\hat{R}$.

$$f_{scale}(\mathbf{R}, A, B, \omega) = w_A + w_B + w_0$$
$$w_A = A \cdot r$$
$$w_B = B \left( \frac{1}{1 + e^{\omega(25 - 50r)}} \right)$$
$$w_0 = \frac{(1 - A - B)}{2}$$
$$Constraints: \quad 0 \leq \{\mathbf{R}_{ij}, A + B, \omega\} \leq 1;$$
(2)

where $A$, $B$ and $\omega$ are the hyper-parameters to be optimized, and $\mathbf{R}_{ij}$ are the values of the input matrix $\mathbf{R}$.

This rescaling function considered is a linear combination of two functions, a linear function, and a sigmoid function, each multiplied by one parameter, $A$ and $B$, respectively. Parameter $A$ weights the linear function, while $B$ weights the sigmoid function. Parameter $\omega$ is the sigmoid parameter that multiplies the input of the rescaling function, centered at 0.5. We consider a constraint in these parameters, in such a way that the sum of $A$ and $B$ cannot exceed 1, to ensure that the output values are bounded between 0 and 1. The rescaling function can take various forms centered at 0.5, with increasing values along the input (positive derivative). The proposed scaling function is flexible enough to represent linear and exponential scales, however,

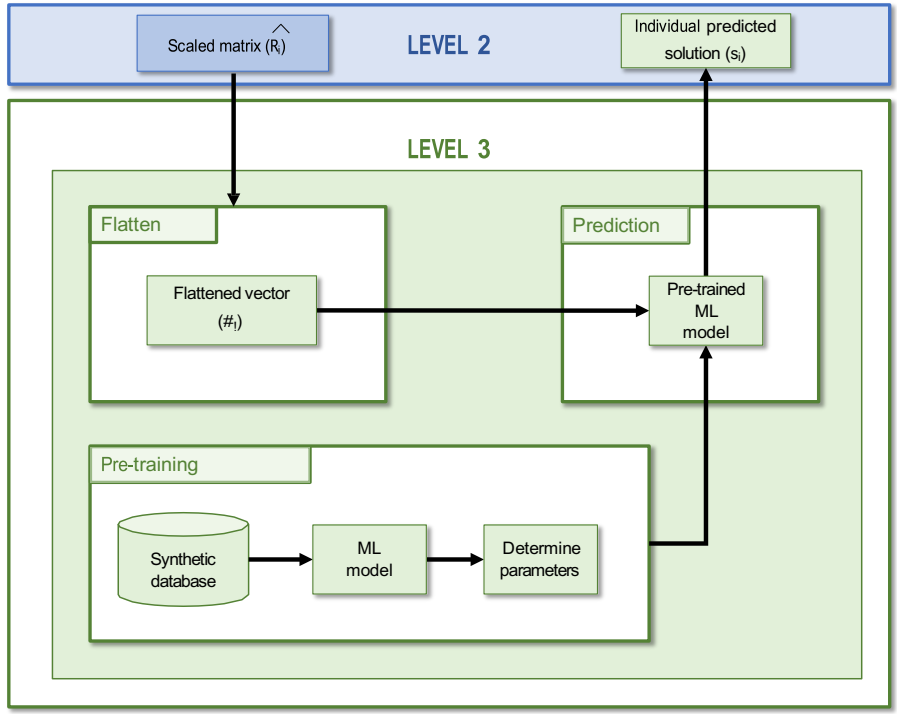

Figure 6: Machine Learning processes involved in the model's level 3: Prediction, Deployment, and Training.

other scaling functions may be considered and/or added.

*2.4. Level 3: Prediction of the segmentation*

The prediction layer is a Level 3 layer (see Figure 1), and its main purpose is to perform a segmentation into an unknown number of groups. After pre-processing the correlation matrices as indicated in the previous sections, we use ML algorithms to detect correlated segments for linear scales. To make the proposed approach completely general, a pre-training of the ML approaches is needed and is done on a synthetic database, generated as a result of modifying several parameters related to the noise present in the system. We show in this research that this pre-training allows the system to learn to solve the task on a pre-defined design scale, without using a specific real problem database.

Figure 6 shows a scheme of this part of the model: we propose to use a synthetic database to train the ML model (this synthetic database is used by any of the proposed ML models). The pre-trained model is then incorporated as a fixed module into the system in Level 3.

In terms of model inference, after being trained with the synthetic database, the input matrices are serialized and flattened to a vector of size of $1 \to T^2$ (except for a few cases such as Convolutional Neural Networks), and a bias term is added. This flattened matrix is then used as input of the pre-trained regressor to generate an output vector, $s_i$, for each input submatrix at Level 3. These output vectors contain values associated with probabilities, as they are real values ranging from 0 to 1. These values indicate the probability that each segment is the beginning of a new group of elements, according to the encoding defined in the specific problem. Values close to 1 represent that there is a high probability that the segment is the first element of a new group, while values close to 0 represent that there is a high probability that the segment belongs to the previous group. Each prediction obtained is transmitted again to Level 2, Layer 2 of the model. For more information on possible algorithms used in Level 3, please refer to Subsection 3.2.

## 2.5. Level 2, Layer 2: Overlap and threshold layer

The overlap and threshold layer is a Level 2 layer (see Figure 1), and its main purpose is to merge the predicted segmentation of the $V$ sub-matrices (each of size $T$) into a unique segmentation vector of size $M_0$. After the pre-trained ML model predicts the indexes of elements belonging to a new group, Level 2, Layer 2 awaits for all matrices to be processed and uses the proposed technique called "Overlap Mean" to merge all predictions from Level 3 into a single prediction ($s_0^\uparrow$). The result is a vector of predictions with real values between 0 and 1, indicating the probability of each index starting a new group of elements. Next, the vector is transformed into a binary vector ($s_0$) using a conventional thresholding technique with parameter $th$. This parameter, along with the parameters of the scaling function, is trained using the metaheuristic characteristic of Level 2, Layer 2, as shown in Figure 5 (right side). The segmentation vector already binarized ($s_0$), is transmitted to Level 1, Layer 2 of the model, to perform a simple transformation on the vector and to be able to present the CoSeNet model's output data with the expected format.

Regarding the "Overlap Mean" (OM) method applied, it is an overlapping method that maintains similarities with the $WOCD$ method presented above. Figure 7 illustrates the $OM$ method applied, where the predictions from Level 3 are first organized, and $OM$ and thresholding are performed on the information of each prediction.

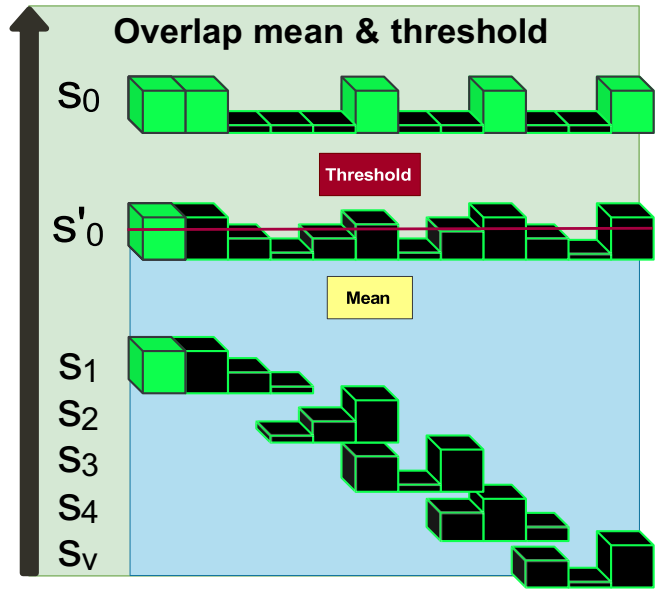

Figure 7: Graphical overview of the Overlap & Threshold process.

## 2.6. Level 1, Layer 2: Reconstruction of the final solution matrix

The reconstruction layer is a Level 1 layer (see Figure 1) in charge of producing a segmentation vector $s_{out}$ sized $1 \rightarrow M_{in}$ (same size as the problem's input matrix size).

After the predictions have been combined into a single prediction at Level 2, Layer 2, the final step is handled by Level 1, Layer 2 of the model, which is responsible for the final formatting and presentation of the segmentation information in a binary output vector, noise-free (See right side of Figure 3). This process involves trimming the segmentation binary vector ($s_0$) using information about the original matrix size, and producing a binary vector of size $1 \rightarrow M_{in}$. The last elements of the vector, which correspond to those added by the $IP$ to make the matrix divisible by the system's throughput, are now removed, to obtain a final binary prediction vector ($s_{out}$).

Finally, the prediction vector can be used to construct a matrix that represents the original binarized correlation matrix, based on the model's predictions. The overall process is illustrated in Figure 8, which provides an overview of all the steps involved in the segmentation prediction, from the input matrices to the final output matrix in a sequential perspective.

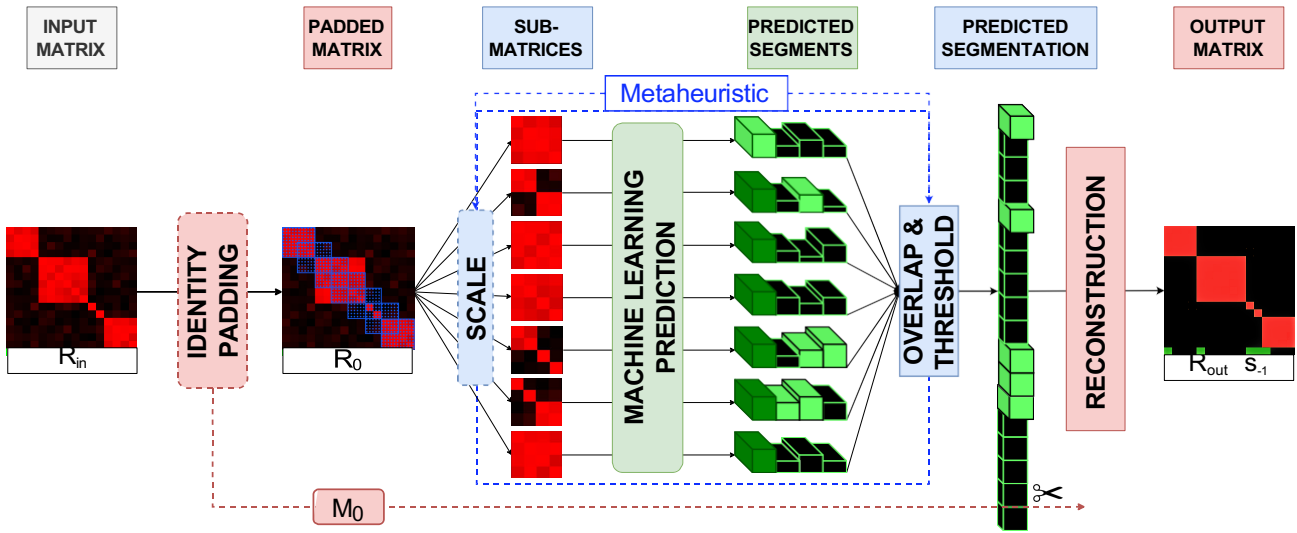

Figure 8: System overview in a sequential perspective for a given machine learning model.

## 3. Evaluation of pre-trained ML prediction models

The proposed CoSeNet approach is capable of solving any matrix segmentation problem, but it must be tuned to improve performance for specific problems. In addition, we can use several different ML methods in the proposed approach, including different forms of linear regression, neural networks, deep algorithms, etc., and we can, of course, carry out comparative tests to check which one behaves best in terms of matrix segmentation performance. These ML methods can be incorporated into the model as pre-trained models, using synthetic databases to tune them. In this section, we evaluate this possibility and analyze the performance of different ML approaches trained on a synthetic database, and how is the transferability of these ML approaches, i.e. the change of performance of these ML models when we apply them to databases with different variables.

### 3.1. Synthetic databases

We have generated several synthetic databases using as inputs the corresponding segmentation vectors (ground truth solutions). Therefore, we have to determine several characteristics: 1) vector's size $M_{in}$, 2) the number of segments $N_g$, and 3) the amount of noise added (determined by the mean and the variance).

The different correlation matrices used in this work have been generated with segmentation vector's sizes $M_{in}$ = 8, 16, and 32, and several segments (groups) and mean and variance of the added noise as presented in Table 1. Each database is then formed by 32,768 ($2^{15}$) pairs of matrix-segmentation, in which we divide 70% to train, 20% as validation set, and 10% to test.

Note that the diagonal of the matrix is finally set to 1 since all correlation values must be maximum on the main diagonal. This process can be repeated as many times as desired to generate a database with specific parameters, resulting in a set of noise-free solutions and a set of noisy data points.

To train some ML models (Level 3) that require a pre-selected fixed input size (throughput $T$), we have generated a synthetic database where matrices sizes ($M_{in}$) match this requirement, therefore, in this case, $M_{in} = M_0 = T$.

Table 1: Databases generated for training the ML model at Level 3 in terms of the matrix input size ($M_{in}$), noise parameters, and cluster metrics.

| $M_{in}$ | Gaussian Noise | | Number of groups ($N_g$) | |
|---|---|---|---|---|
| | Mean | Variance | Mean | Variance |
| 8 | 0.00 | 0.0 | 3.0 | 1.0 |
| 8 | 0.01 | 0.1 | 3.0 | 1.0 |
| 8 | 0.01 | 0.2 | 3.0 | 1.0 |
| 8 | 0.01 | 0.3 | 3.0 | 1.0 |
| 8 | 0.01 | 0.4 | 3.0 | 1.0 |
| 8 | 0.02 | 0.5 | 3.0 | 1.0 |
| 16 | 0.00 | 0.0 | 4.0 | 2.0 |
| 16 | 0.01 | 0.1 | 4.0 | 2.0 |
| 16 | 0.01 | 0.2 | 4.0 | 2.0 |
| 16 | 0.01 | 0.3 | 4.0 | 2.0 |
| 16 | 0.01 | 0.4 | 4.0 | 2.0 |
| 16 | 0.02 | 0.5 | 4.0 | 2.0 |
| 32 | 0.00 | 0.0 | 8.0 | 2.0 |
| 32 | 0.01 | 0.1 | 8.0 | 2.0 |
| 32 | 0.01 | 0.2 | 8.0 | 2.0 |
| 32 | 0.01 | 0.3 | 8.0 | 2.0 |
| 32 | 0.01 | 0.4 | 8.0 | 2.0 |
| 32 | 0.02 | 0.5 | 8.0 | 2.0 |

### 3.2. Machine-Learning models considered

Supervised ML algorithms are those methods that use a labeled dataset of input-output pairs $D = \{(\mathbf{x}^i, y^i) | 1 \leq i \leq n\}$ to infer the general relation $y = f(\mathbf{x})$ between the input variables $\mathbf{x} = (x_1, x_2, \cdots, x_n)$ (also called predictors) and the output variable $y$. Except for the non-parametric methods which directly use the input-output pairs of the database $D$ for providing predictions, most of the supervised ML methods obtain the input-output map $f$ by minimizing a *loss function* $L$, which penalizes a kind of error, into a specific parametric function space $f \in F = \{f(\cdot, \varepsilon) | \varepsilon \in \mathbb{R}\}$, shown in Equation (3):

$$f^* = \arg\min_{\varepsilon \in \mathbb{R}} L(f(x, \varepsilon), y)) \qquad (3)$$

Different parametric function spaces $F$ together with their learning algorithms drive a huge variety of methods. We consider three categories of them: QP-based (quadratic programming), ensemble, and backpropagation-based methods, analyzed in Sections 3.2.1, 3.2.2 and 3.2.3, respectively. Table 2 shows the acronyms of the methods used in this manuscript.

### 3.2.1. QP-based

Quadratic programming-based methods, or simply QP-based, are those supervised ML algorithms that involve a quadratic, $L_2$ or least-squares minimization problem in their loss function, shown in Equation (4):

$$f^* = \arg\min_{\varepsilon \in \mathbb{R}} \|f(x, \varepsilon) - y\|_2 + \vartheta L_\varepsilon(\varepsilon) \qquad (4)$$

with possibly other regularization terms $L_\varepsilon(\varepsilon)$ weighted by a parameter ($\vartheta$).

The Linear Regression method (LR) assumes a linear relationship between the inputs variables and the output, as shown in Equation (5):

$$y = w_1 x_1 + w_2 x_2 + \cdots + w_n x_n + \varpi \qquad (5)$$

| Method or algorithm | Acronym |
|---|---|
| Adaboost Regressor | AR |
| Bayesian Ridge | BR |
| Convolutional Neural Network | CNN |
| Decision Tree Regresor | DTR |
| ElasticNet | EN |
| Extra Tree Regressor | ETR |
| Extra Trees Regressor | ETsR |
| Linear Regressor | LR |
| Linear SVR | LSVR |
| Lasso Regressor | Lasso |
| Lasso Lars | LL |
| Multi-Layer Perceptron | MLP |
| Orthogonal Matching Pursuit | OMP |
| Random Forest Regressor | RFR |
| Ridge Regressor | Ridge |
| Random Trees Regressor | RTsR |
| Singular Value Decomposition | SVD |
| Support Vector Regressor | SVR |
| Deep Clustering | DC |
| Hierarchical Clustering | HC |
| Louvain's method | Louvain |
| Modularity Maximization | MM |
| SegCorr algorithm | SegCorr |
| Spectral Clustering | SC |
| Genetic Algorithm Optimization | Genetic |
| Particle Swarm Optimization | PSO |

Table 2: Alias and acronyms of the Machine Learning methods and algorithms considered in this work.

and exclusively minimizes the mean square error. Such a minimization problem directly yields a closed-form solution through the pseudo-inverse. Other linear regressors also include a regularization term in the loss function, such as Lasso [29] (Lasso), Ridge [30] (Ridge) and ElasticNet [31] (EN) that additionally minimize the $L_1$ norm ($\|\varepsilon\|_1$), the $L_2$ norm ($\|\varepsilon\|_2$) and a linear combination of both norms ($\omega_1 \|\varepsilon\|_1 + \omega_2 \|\varepsilon\|_2$) of the coe"cients $\varepsilon$, respectively. LassoLars [32] (LL) fits also a linear regressor that involves the $L_1$ norm as a regularization term, similar to Lasso, but least-angle regression (LARS) obtains the best variables following the equiangular direction. Bayesian Ridge [33, 34] (BR) is also a linear regressor method that includes Ridge regularization. However, BR presumes a prior distribution of the $\vartheta$ parameter that fits in consecutive iterations with the training set. This method also estimates the best weight $\vartheta$ using the Bayesian method.

Orthogonal Matching Pursuit [35, 36] (OMP) is a sparse approximation method that finds the so-called best matching projections of the data onto the codomain of an over-complete linear function, represented by a matrix $D$ (frequently named as *dictionary*).

The objective is to represent the input-output function ($f$) from the Hilbert space $H$ as a finite linear combination ($\hat{f}_N$) of $g_{\vartheta_n}$ (known as atoms) extracted from columns of $D$, as follows in Equation (6):

$$\hat{f}_N(t) = \sum_{n=1}^{N} a_n g_{\vartheta_n}(t). \qquad (6)$$

Matching Pursuit seeks the atoms, one at a time, to maximally reduce the approximation error, that is, choosing the atom with the highest inner product with the function $f$ subtracting the approximation that uses only that one atom.

Support Vector Regression (SVR) [37, 38, 39] is a well-established algorithm for regression and function approximation problems. The SVR formulation is quite similar to its classification counterpart. It also establishes an optimization problem where few support vectors are found to approximate the regressor. Besides, it is common to adopt an appropriate non-linear mapping $\varrho : R^n \to R^p$ that transforms samples to a higher-dimension feature space $R^p$ ($n \ll p$). By solving the dual optimization problem in the feature space $R^p$, the scalar product $K(x_i, x_j)$ of the high-order space $R^p$ is called *kernel* [39]. This *kernel trick* has been used for a large number of problems and applications in science and engineering [40], especially the linear, polynomial, or Gaussian kernels. In this article we use the linear (LSVR) and the Gaussian (SVR) kernels. Details on the solution process for the SVR algorithm and its tuning and optimization can be found in [39].

Singular Value Decomposition (SVD) [41] is a matrix factorization method and a generalization of eigen-decomposition of squared matrices. Consider that $X \in \mathcal{M}_{m,n}(R)$ is a $m \times n$ matrix in the real field. We know that the matrix $X^T X$ is a positive semi-definite symmetric squared matrix. As a consequence of the spectral Theorem and Sylvester's Theorem [41], matrix $X^T X$ is a diagonalizable matrix whose eigenvalues are all non-negative real values $\{\vartheta_i \geq 0\}$. Ordering these eigenvalues $\{\vartheta_1 \geq \vartheta_2 \geq \cdots \vartheta_n \geq 0\}$, the $\varsigma_i = \sqrt{\vartheta_i}$ is called the $i$th singular value of the matrix $X$. Besides, the factorization of the matrix $X$ can be expressed as $X = U S V^T$, where $U \in \mathcal{M}_{m,m}(R)$, $V \in \mathcal{M}_{n,n}(R)$ are orthogonal matrices and the matrix $S \in \mathcal{M}_{n,n}(R)$ is build with the singular values of $X$, in descending order, in its principal diagonal, called the *singular value decomposition* of the matrix $X$.

*3.2.2. Ensembles*

Ensemble methods improve the predictive performance of single ML models, based on combinations of different training models. They assume that the contribution of several base ML models, named *learners*, can enhance the prediction ability and even overcome the robustness and generalization capacity of more complex ML methods [42]. Usually, learners are simple ML methods, especially decision trees or linear regressors.

Decision Tree Regressors [43, 44] (DTR) are ML methods that build a tree graph that follows branching decision paths through the data to provide a prediction. In decision trees for regression, the threshold parameters that split the data are calculated from the whole training data set following a specific criterion, such as looking for the best gain of information possible in the current node. Extremely randomized trees [45] (ETR) differ from classic decision trees in the way they are built that choose the best split among a randomly selected set of features. Decision trees are not ensemble methods themselves but are used in ensemble methods as learners. This is the case of extra tree regressors, random forest, or AdaBoost, we briefly comment.

Random Forest [46] (RFR) is the most renowned bagging-like technique for both classification and regression problems. It specifically uses regression trees as learners and differs from the pure bagging technique in that the topology of the trees varies among them. Trees of the ensemble, the forest, may have different lengths or topology, or use different input variables which greatly increase the variability of the learners. Its main advantage lies in its generalization capacity, achieved by compensating the errors obtained from the predictions of the different regression trees. Once the regression trees have been generated, and each has obtained its prediction, an averaging scheme is taken into account for the final prediction [46]. Extremely randomized trees [45] (ETsR) have an extra level of randomness compared with random forests. After choosing a random subset of candidate features for each node, thresholds are drawn at random for each candidate's feature instead of looking for the most discriminative thresholds as a random forest does. Then, the best of these randomly-generated thresholds is selected.

Adaptive Boosting (AB) [47] is the widest-used boosting technique in the history of ensemble learning. As with all boosting methods, AdaBoost proposes to train the learners sequentially, in such a way that each new learner requires that the previous learner had been trained before. Each base learner has the same topology in the queue and focuses on the data that was mispredicted by its predecessor, to iteratively adapt its parameters and achieve better results. In this way, learners are dependent on them. In boosting, all the learners use the whole training dataset for computing their parameters, i.e., there is no bootstrap sample step.

*3.2.3. Back-propagation-based*

Back-propagation-based methods are those supervised ML methods that allow the application of any training iterative method, such as the gradient descent method, taking advantage of the sequential architecture. A sequential topology of the architectures allows for reducing the number of calculations in each iteration. Almost all feed-forward Artificial Neural Networks (ANNs) are back-propagation-based methods. In this work, we detail both the multi-layer perceptron (MLP) and the Convolutional Neural Network (CNN).

An MLP [48] is a kind of ANN which has been successfully used in classification and regression problems. The MLP is a feed-forward network composed of an input layer, several hidden layers, and a final output layer, all sequentially placed. Each layer is composed of a collection of neurons that are connected to the neurons of the consecutive layer through weighted links. The weight values are calculated from a su"ciently large database of input-output pairs minimizing the error between the output given by the MLP and the corresponding expected output in the training set. The number of hidden layers and their neurons are also parameters to be optimized [49, 50].

Convolutional Neural Network (CNN) [51] is also a kind of ANN that has been extensively applied to computer vision. As MLPs, they are feed-forward networks and are composed of an input layer, several hidden layers, and a final output layer, all sequentially placed. However, contrary to MLPs, each layer is not fully connected to its predecessor. Layers calculate the mathematics operation of convolution through several kernels that characterize them and with the output data of the previous layer. Training CNNs usually requires a huge amount of data since it is common to place several CNN layers sequentially and the number of parameters quickly grows.

*3.3. ML algorithms comparison*

In this section, we evaluate the results obtained on the synthetic database described in Section 3.1 using the di!erent ML algorithms considered in Subsection 3.2 as pre-trained models (in Level 3). Note that we are comparing here ML models that specifically solve the segmentation problem, without considering the formatting and scale part of the system, which is appropriate for tuning real problems.

We use several metrics to model the segmentation error. Over the defined encoding, the segmentation vector is a binary vector with the size of the system's throughput. Thus, the first metric is the *MSE*, i.e., the Mean Square Error of the prediction, and measures the distance between the expected output and the predictor output on a quadratic scale. The second is the *MAE*, i.e. Mean Absolute error. It measures the distance between the expected output and the input on a linear scale. We also use the R2 metric (also known as the coe"cient of determination or Pearson's coe"cient squared), which is a metric that measures the proportion of the variance in the dependent variable that is explained by the independent variables. Finally, we also make use of WindowDi! (*WD*) [52], which measures the proportion of frames for which the predicted and actual outputs di!er. It is a popular evaluation metric for segmentation problems and is designed to obtain segmentation metrics in the same coding in which we have included this problem. This metric is widely used in segmentation problems, especially, in text and topic segmentation [53].

Regarding the implementation of the ML methods described above, we have used the Scikit-learn (Sklearn) toolkit for Python [54] to implement the majority of the ML algorithms in this comparison, except for CNN, MLP, and SVD algorithms. Sklearn is an open-source library that provides a comprehensive set of algorithms for classification, regression, clustering, and dimensionality reduction. CNN and MLP algorithms were implemented using TensorFlow [55] and Keras [56], while SVD was implemented using NumPy [57].

Figure 9(a) shows the comparison of the proposed ML models' performance on the synthetic database, taking into account *MSE*, *MAE*, R-squared (*R2* or $R^2$, note that ↓ *R2* has been represented) and *WD* for the whole range of noise variances considered, while 9(b) shows the average model's performance for all variances considered.

We can observe that some models (Lasso, ElasticNet, and LassoLars) do not seem to converge, or are not adequate to solve the problem. The models that achieve the best performance, for all metrics considered, seem to be SVR, MLP, and Ridge regression. Both Backpropagation and QP-based models solve the problem, while Ensemble-based models do not seem to achieve high performance, with AdaBoostRegressor being the best of them. Note that CNN can solve the problem, but does not achieve the best segmentation metric (*WD*), although *MSE* and *MAE* are above most regressors.

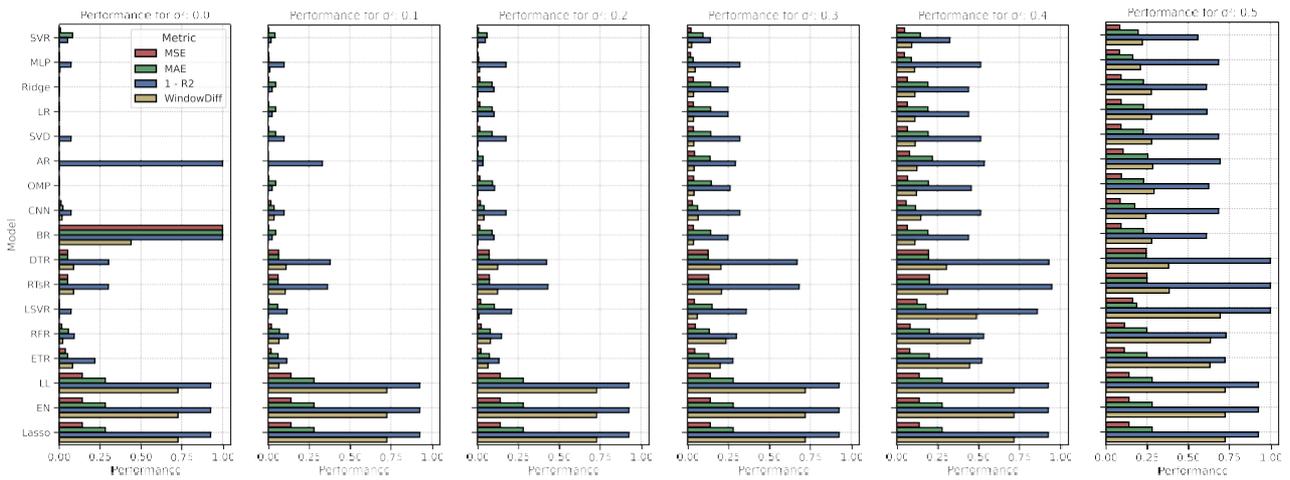

(a) Model performance comparison for each selected variance $\varpi^2$.

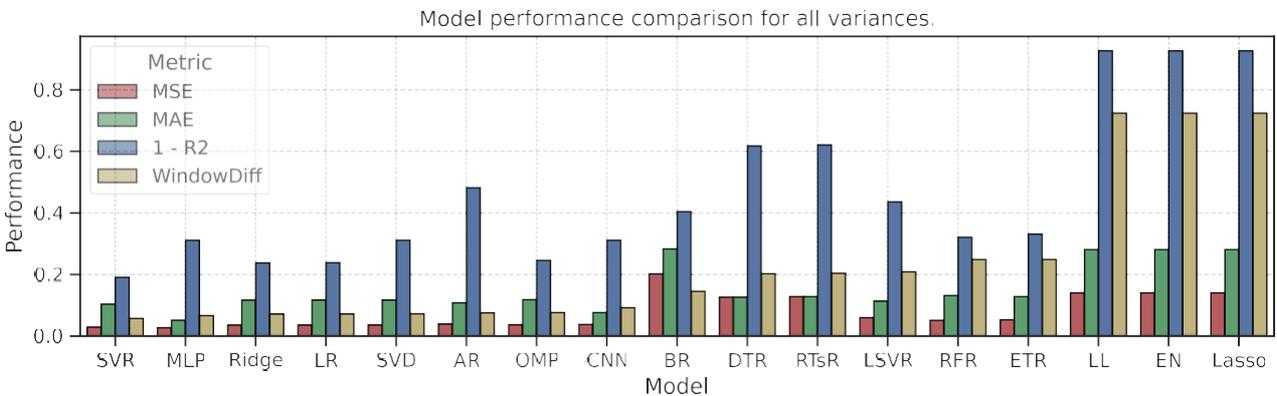

(b) Model performance comparison for all the selected variances $\varpi^2$.

Figure 9: Models' performance comparison when no data standardization is considered for each variance (a) and all variances (b).

Regarding the behavior of the models as a function of noise variance, we can see that all metrics perform worse when the noise variance is increased, as expected. However, it is interesting to note that Bayesian Ridge is not able to converge if there is no noise in the training matrices. Moreover, the Backpropagation-based models do not achieve an outstanding score for the noise-free case, but it is the simplest linear regressor that can solve this case, achieving perfect scores and proving that this is a problem that is possible to solve with a multilinear model. When the input matrices are noisy, multilinear models can solve the problem, however, they are not the best (although feasible) solutions in terms of performance.

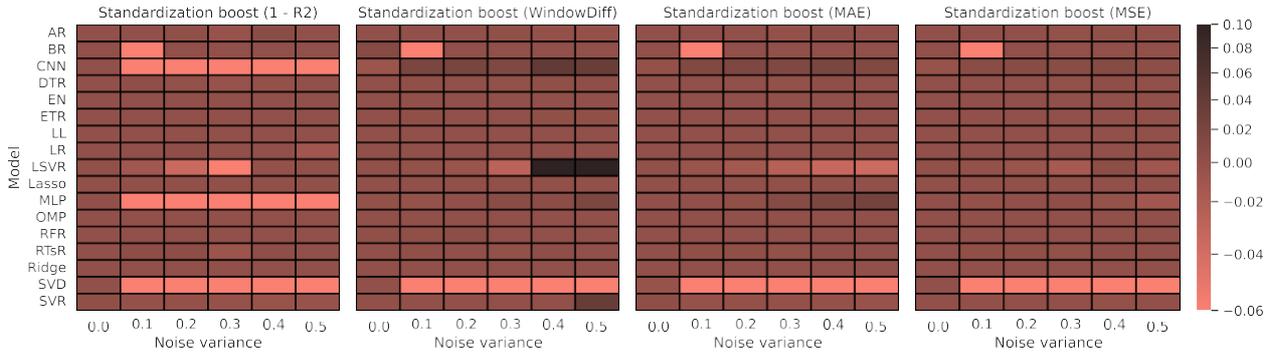

Figure 10: Performance improvement (boost) of the models for each metric as a function of the standardization of the input data and noise variance.

For further testing the ML models considered, we have performed the same experiment with a standardization (or normalization) of the data [58]. Within the same model, we train a classical standardizer that subtracts the mean of the training data values and divides it by the standard deviation. This may improve the performance of some of the regressors, but may also reduce the performance of others. Figure 10 shows the performance improvement if the standardizer is added to the proposed approach. We can see that for most models it is not critical. However, there is a slight improvement for models that achieve very good performance (SVR, MLP, and CNN), so it is advisable to introduce the *Standardizer* for these cases. On the contrary, note that for the LSVR, the segmentation (*WD*) improves and the *MAE* worsens, so we must be careful if we decide to implement this regressor with data standardization. Also, for models such as SVD and one of the BR cases, the *MAE* worsens, so the standardizer is not recommended.

Given the results, we can conclude that the best-performing and most stable models are Support Vector Regression, MultiLayer Perceptron, and Ridge regression. Ridge is the best model in the less noisy cases, SVR is the best ML model in the noisiest cases, and MLP results in an intermediate solution of both previous cases.

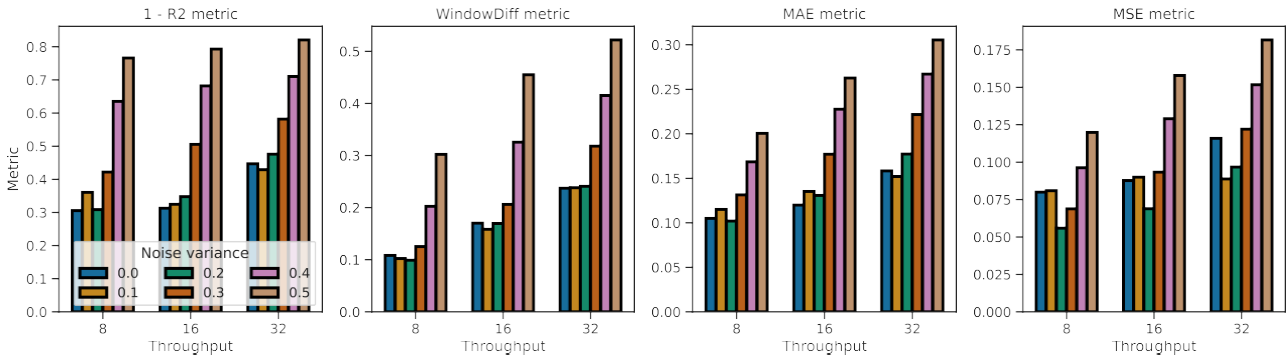

Figure 11: Error behavior as a function of variance and system throughput for values of 8, 16, and 32.

Regarding the computation time for each ML algorithm, note that the SVR takes a long time to train, especially for cases where noise variance and throughput are high. On the other hand, MLP does not take such an excessive time to train, and Ridge regression computation time is extremely low. This is also extrapolated to the processing time. Ridge is the fastest of all methods compared, while MLP is an intermediate solution between SVR and Ridge, also in terms of the algorithm's complexity. In addition, we performed an analysis of the model's behavior as a function of the throughput. Figure 11 shows that the performance metrics worsen linearly with model throughput; and that they increase exponentially with noise variance for all cases. The noise-free case is an outlier since many of the models have di"culties converging to a solution. Thus, it is su"cient to select a model that does not exhibit instabilities. It is important to choose the right throughput for the problem to be solved since we can slightly worsen the performance of the system if we choose a higher throughput or we may be losing information about the data, and we may be performing too many calculations for a too-small throughput. After a first analysis, we have seen that a throughput $T = 16$ adequately solves most problems without increasing the error too much.

Consequently, from these experiments on synthetic databases, we establish that the best regressors for this problem are the SVR with Standardization, MLP with Standardization, and Ridge regression without Standardization. All three models are candidates to be implemented in the approach as pre-trained models for segment prediction and SVR is the best one in terms of performance.

*3.4. Transferability study*

Once we have selected the models that best fit in the pre-training stage (SVR, MLP, and Ridge), we can assess performance when using a pre-trained ML model for a given throughput and noise variance. We call this parameter the *Transferability* of the model. Let us define the metric Transferability ($\mathcal{T}$), in Equation (7), as the average of the performance, given a metric $m$ (*MAE*, *MSE*, *WD*, etc.), for all parameters $p$ considered (i.e. $\varsigma^2$). Note that the lower the Transferability, the better the performance will be for di!erent input databases.

$$\mathcal{T}_p^m = \frac{1}{N_p} \sum_{i=1}^{N_p} m_{i,p} \tag{7}$$

where $N_p$ is the length of the set of parameters for a specific metric.

In this work, we evaluate the Transferability for two metrics ($m = \{MSE, WD\}$) and six possible parameters ($p = \varsigma^2 = \{0, 0.1, 0.2, 0.3, 0.4, 0.5\}$, from no-noise to a variance of 50% of the range of values), considering three di!erent throughput scenarios ($T = \{8, 16, 32\}$), and the results are presented in Figure 12 for the three best-performing ML models (a) Ridge, (b) MLP, and (c) SVR regressors. The solid blue line represents $\mathcal{T}_{\varpi_i^2}^{MSE}$ while the solid orange line represents $\mathcal{T}_{\varpi_i^2}^{WD}$. The ML model is trained using the training dataset for one given $\varsigma_i^2$ and tested for all test sets for all possible $\varsigma_i^2$, obtaining the average Transferability. Additionally, each ML model has been trained using the training dataset obtained by concatenating all variances' train subsets and tested using the test dataset obtained by concatenating all variances' test subsets. These results are presented with dashed blue and red lines for the *MSE* and *WD* respectively. Several conclusions can be obtained from these results. First, the Transferability (for both *WD* and *MSE*) improves as the ML model is trained with higher noise variance. Second, the Transferability when training with a database that contains all possible noise variances is worse than that of the model trained with higher noise (solid lines are below dashed lines as the noise increases). This can be explained, for example, for some techniques such as neural network regularization, as a model generalization is improved by adding Gaussian noise [59].

Since our synthetic databases are very large and randomly generated, it is relatively frequent to find repeated data, especially for small throughput values, therefore adding noise can reduce overfitting and improve Transferability and generalization.

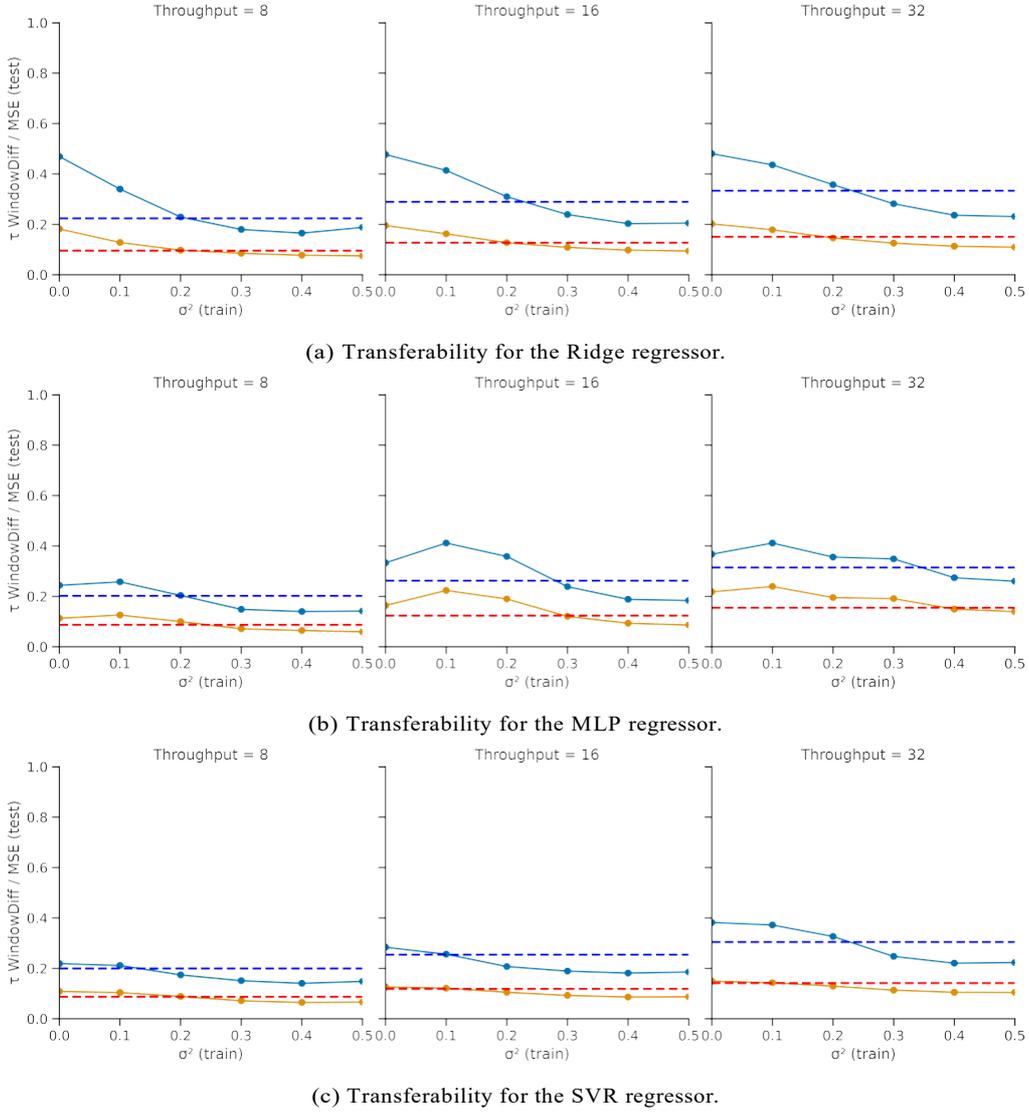

Figure 12: Transferability performance for the best ML prediction models for (a) Ridge, (b) MLP, (c) SVR regressor, and different Throughputs. Solid orange lines represent $T^{WD}_{\omega^2_i}$ and solid blue lines represent $T^{MSE}_{\omega^2_i}$ for $\omega^2_i = \{0.0, 0.1, 0.2, 0.3, 0.4, 0.5\}$. Dashed red and blue lines present the results when the models are trained with a training dataset containing all noise variances, for metrics *WindowDiff* (*WD*) and *MSE*, respectively.

Finally, we analyze memory space occupied, performance (taking as reference the segmentation metric), and speed at which the models can obtain the segmentation from correlation matrices. Figure 13 is a radar chart showing the statistics of each model as a function of the selected metrics. Note that a square root scale has been used to represent and differentiate metrics among the algorithms, as numerical results are very close (see Figure 9).

The most balanced of the three models is the MLP, while the best-performing is the SVR. The one that achieves the highest speed and the smallest memory footprint is Ridge regression, which is the most useful option if we want to optimize computational load and the one that can have the greatest impact in a real deployment. On the contrary, the SVR is a restrictive option due to its excessive computation time. Summarizing the results, SVR takes 43.4 ms on average to compute each of the input matrices, while MLP takes 19.41 $\mu$s and Ridge 1.89 $\mu$s per matrix. In terms of memory, SVR loads between 313.6 kB and 4.2 GB, MLP between 120.2 kB and 21.0 MB, and Ridge between 2.5 kB and 131.7 kB, depending on the throughput. Finally, we can conclude that there is no optimal regressor and that we can choose between these three solutions depending on the problem and system requirements.

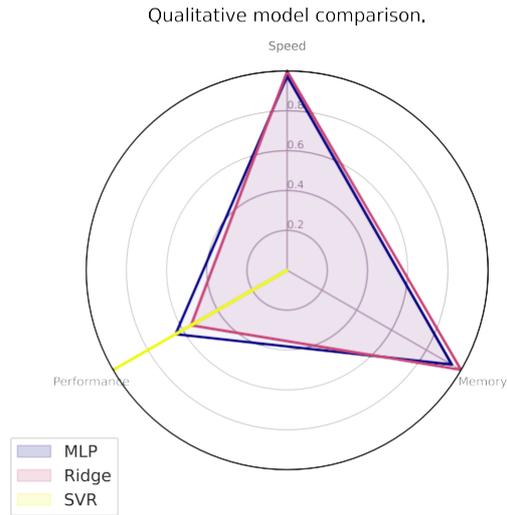

Figure 13: Qualitative comparison (using a squared-root scale) between MLP, Ridge, and SVR regressors in terms of computational speed, memory usage, and *WindowDiff* performance.

## 4. Experiments over a real NLP problem database

In this section, we analyze the performance of the proposed approach in a real-life problem in the field of Natural Language Processing (NLP). Specifically, the problem at hand consists of segmenting a given text into different topics or stories (known as *Text Segmentation* in NLP). To generate the input text, we extract information from a random Wikipedia article and divide the text into sentences, to construct each section/group of the input text. Therefore, we will know that all sentences in this article are related to each other and form a group. To ensure solid stories, we require that the articles contain a minimum of 40 words. We keep on adding groups/sections to the input text until the total number of sentences is equal to $M_{in}$. Thus, the number of sections in the input text varies depending on the length of each article randomly chosen.

Once we have the input text, we obtain the correlation matrix (input to our proposed model) using BERT [28] (a pre-trained language model that uses Deep Learning Transformers [60]), by obtaining a similarity value between each pair of sentences in the text. These similarity values can be arranged as a square correlation matrix (size $M_{in} \times M_{in}$). Note that the data in these correlation matrices match the requirement of spatial ordering, as the sentences of a story are always spatially ordered and appear in sequence.

The model proposed in this work will have to determine that each Wikipedia article corresponds to a different segmented group. Failing the segmentation if several groups are found within one article or if no group is found at the beginning of one article.

Therefore, the Ground Truth segmentation vector is built at the time of extracting the articles of the Wikipedia text, using the sentence number where a story starts.

Table 3 shows the information regarding the correlation matrices constructed using Wikipedia articles. These correlation matrices are later divided, and 70% of the samples are dedicated to training the model, 20% to model parameters' validation, and 10% to testing.

Note that the input matrix size ($M_{in}$) not necessarily matches the system's throughput ($T$). Any other correlation matrix size where $M_{in} \leq 256$ can be constructed using any of the subsets presented in Table 3.

In this case, we have chosen to implement Ridge regressor as the pre-trained ML approach (trained on the synthetic database explained in Subsection 3.1) that is fed with this problem's specific database (correlation matrices obtained from Wikipedia articles). It should be noted that speed is a crucial factor in our approach for real applications since it includes a heuristic optimization of the model, that would take too long if a slow ML model for training is chosen. Specifically, we optimize five hyperparameters: three for the scaling function, one for the threshold value, and one for the system's throughput ($T = 8$, 16, or 32). This hyperparameters' optimization has been implemented using a Particle Swarm Optimization (PSO) [61] algorithm and a Genetic Algorithm (GA) [62]. The methodology we use to optimize the hyperparameters of the model is the following: we first run the PSO algorithm and obtain the best 5 individuals based on their fitness scores. Similarly, we ran the GA algorithm and obtain the best 5 individuals. Then, we use the validation dataset to assess the performance of the 10 best individuals and select the best candidate.

Table 3: Some example correlation matrices generated for testing the ML model in terms of the matrix input size ($M_{in}$) and the way they are generated.

| $M_{in}$ | Number of correlation matrices | Generator |
|---|---|---|
| 256 | 4,152 | Wikipedia + BERT generated |
| 128 | 8,304 | Subset of $M = 256$ |
| 64 | 16,608 | Subset of $M = 128$ |
| 32 | 33,216 | Subset of $M = 64$ |
| 16 | 66,432 | Subset of $M = 32$ |
| 8 | 132,864 | Subset of $M = 16$ |

Table 4 shows the 5 best sets of parameters obtained for each algorithm. The results show that the GA algorithm outperforms the PSO algorithm in finding the optimal parameter combination. The best candidate obtained from the GA algorithm had a higher *WD* performance and lower throughput compared to the best candidates obtained from the PSO algorithm. The parameters of the simulation were 20 epoch/iterations and 200 individuals for the GA, computing 100 new individuals per epoch. The crossover rate and the mutation variance were 0.5 and 0.1 respectively, with a uniform crossover mask. Regarding the PSO algorithm, the cognition and social factor were set up to 1, and the inertia was set to 0.5 for 30 particles. The best result was obtained for the best individual of the Genetic algorithm with a throughput $T = 32$. The best combination of parameters for the Wikipedia database suggests a 34% of linearity and 36% of non-linearity (for parameters *A* and *B*, respectively) in the re-scale function. The sigmoid variance parameter $\omega$ took a wide range of values, while the threshold value tends to be above 0.5 for all good candidates.

Figure 14 shows an example realization of 14 correlation matrices for a size $30 \times 30$ on the Wikipedia database. The lowest part of the figure displays the successful predictions in green bands (True Positives), indicating the correct detection of segment boundaries by our four-layered algorithm. In contrast, the red bands indicate a new group of correlated sentences that were not detected by the algorithm (False Negatives). The blue bands represent predictions of new element groups when in fact they were not present in the correlation matrix (False Positives). The black background denotes True Negatives, where no new segment is present in the matrix. This figure demonstrates the ability of our approach to accurately identify segment boundaries in correlation matrices, as evidenced by the high number of True Positive predictions. While some False Negatives and False Positives are present, they are relatively small in number and do not significantly impact the overall accuracy of the algorithm. We also realize that most failures occur at the edges of the matrices since these areas present the least context information about the predecessor or successor correlation values of the groups.

*4.1. Performance comparison with state-of-the-art algorithms*

In this section, we compare our approach's performance with several methods proposed in the state-of-the-art, which are used to solve the segmentation problem in correlation matrices.

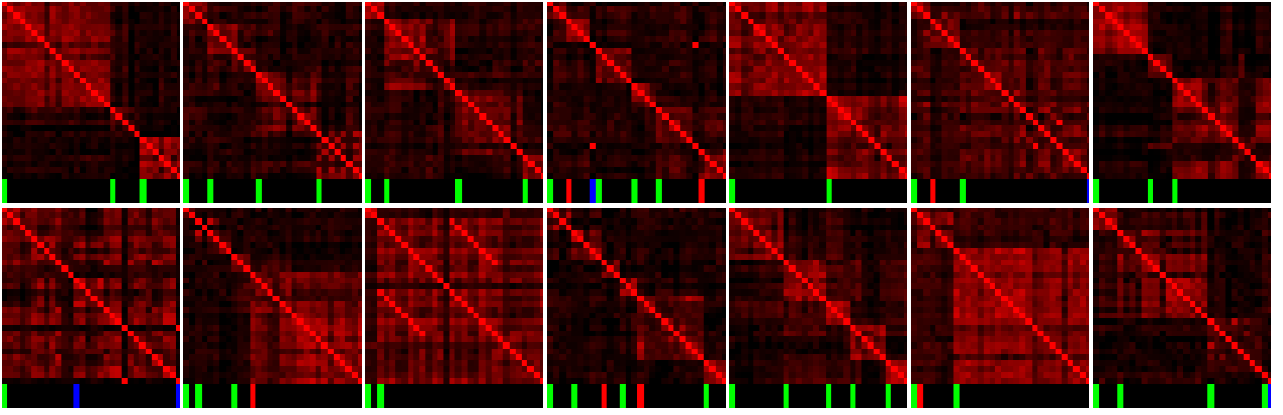

Figure 14: Example Realization of 14 matrices (size 30⨯30) obtained from the Wikipedia Database. Green bands in segmentation denote true positives, blue bands denote false positives, and red bands denote false negatives.

| Ranking | Algorithm | WD (%) | A | B | ω | th | Throughput |
|---|---|---|---|---|---|---|---|
| **1st** | **Genetic** | **19.52** | **0.34046** | **0.36040** | **0.17642** | **0.67790** | **16** |
| 2nd | PSO | 20.48 | 0.55448 | 0.39009 | 0.45960 | 0.93589 | 32 |
| 3rd | Genetic | 20.50 | 0.63856 | 0.00000 | 0.00000 | 0.59944 | 16 |
| 4th | Genetic | 20.51 | 0.76527 | 0.00000 | 0.00000 | 0.67117 | 16 |
| 5th | PSO | 20.52 | 0.55573 | 0.38922 | 0.46114 | 0.93540 | 32 |
| 6th | Genetic | 20.60 | 0.96621 | 0.00000 | 0.00000 | 0.81306 | 16 |
| 7th | Genetic | 20.69 | 0.58012 | 0.26236 | 0.53721 | 0.79231 | 16 |
| 8th | PSO | 51.78 | 0.83080 | 0.38923 | 0.46116 | 0.93653 | 32 |
| 9th | PSO | 53.99 | 0.82296 | 0.37121 | 0.46426 | 0.90916 | 32 |
| 10th | PSO | 54.80 | 0.84770 | 0.36758 | 0.49800 | 0.91590 | 32 |

Table 4: Ranked results for the best 5 individuals of the Genetic and Particle Swarm Optimization (PSO) algorithms for the validation dataset ordered by performance over WindowDiff (WD) considering 3 different possible throughput sizes ($T$ = 8, 16 and 32).

Specifically, we test three types of algorithms: 1) The first one, based on Community Detection (CD) techniques, interprets the correlation matrix as a weight matrix and constructs a graph accordingly. Within this first type, we implement *Louvain's method* [18] and *Modularity Maximization* [27], which are unsupervised algorithms that require a threshold optimization to converge. 2) We have also considered some unsupervised clustering methods, (*Correlation Clustering* ). Specifically, we implement *Hierarchical* clustering [26] and *Spectral* clustering [23] algorithms, the disadvantage of these unsupervised methods is that they require the number of clusters as a parameter (which is unknown sometimes). In addition, we have added here a heuristic approach [8] specifically designed to solve the PBMM problem in an unsupervised way, although it requires the optimization of 3 hyperparameters, and we have also added a comparison with the algorithm proposed in [21] (SegCorr), which is an implementation of a statistical procedure for the detection of genomic regions of correlated expression, used in gene expression, with only 1 hyperparameter. 3) Finally, we have considered *Deep Clustering* based methods, in which a training process is required, but no hyperparameter optimization. These methods use autoencoders, which we have implemented based on CNN (*CNN-DC* ) or MLP (*MLP-DC* ). The way these models cluster the elements is through a 35kB MLP classifier. For all the algorithms used in the comparison, a hyperparameter optimization based on a Grid Search [63] has been performed with a small portion of the database, so that in case the algorithm parameters are critical to the performance of the algorithm they are adjusted properly to the database.

Table 5 shows the results obtained using the synthetic databases described in Section 3, and in the Wikipedia database, for all the algorithms compared. The results shown in this table are the average of the test sets of the synthetic matrices and Wikipedia database. Note that the graph-based (Community Detection) algorithms failed to deliver acceptable results, even after thresholding and transforming weight values. When the correlation matrix noise was high, the only algorithm that performed well in the synthetic database was Modularity Maximization, achieving a 76.12% rate. Deep clustering-based algorithms performed below average in general, without good results either in the synthetic matrices. On the other hand, unsupervised algorithms mainly produced satisfactory results, with performances around 60% for the Wikipedia database and 80% in the synthetic matrices. SegCorr obtained consistent results, higher than other unsupervised methods, with 60.5% of performance in the Wikipedia problem. Finally, the proposed approach with Ridge regression implemented in the prediction layer, showed the best performance for this NLP problem, with a significant 82.60% of performance for the Wikipedia database, and 87.19% for the synthetic database. Note that the difference between the results in synthetic and real problems is small in the proposed approach, though Ridge was trained on the synthetic database. This means that the proposed CoSeNet model is not only robust to noise variation (transferability) but also to scale change.

In terms of execution time, Table 5 shows the computation time (in seconds) of each of the algorithms considered when processing 7500 matrices (7th column: "Comp. time (s./7.5 kS)" that stands for "Computation time (seconds per 7.5 kilo-Samples)"). Note that The smaller this record is, the higher the processing speed of the algorithms. The HC and PBMM algorithms were the fastest (but not the most accurate) approaches, taking 610ms (0.61s) for every 7500 matrices for HC, and 210ms (0.21s) for PBMM. Recall that both are recursive algorithms that perform fast linear operations, making them the fastest among all algorithms analyzed. Ridge regression in the proposed problem achieved the best time records among the pre-trained models, reaching 7500 matrices every 3.13 seconds (2400 matrices per second), as we selected the fastest and lightest in-memory pre-trained model. Other algorithms such as *SegCorr* and *Spectral Clustering* are slower, computing around 300-410 matrices per second. The memory size of unsupervised and graph-based algorithms was zero, as only knowledge of the optimal parameters was required. The memory size of the classification model for Deep Clustering was 35.05kB. The memory size of the pre-trained Ridge-based model was 131.7kB, which allows flexible implementations for deployment.

## 5. Conclusions

In this work, we have proposed a novel approach (CoSeNet) for solving problems related to the segmentation of correlation matrices. The CoSeNet approach uses Machine Learning (ML) algorithms to predict the points that separate groups of elements in correlation matrices. However, the main problem is that these models usually require fixed correlation matrix input sizes. This challenge is solved in the proposed architecture with a first "formatting" level. The second problem found is that each function used in a different problem to quantify the correlation between elements has different properties, therefore the scale of the input matrix is different. This challenge is solved by "scaling" the values (at the second level of the proposed architecture) to have a pre-trained and ready-to-use model. Finally, in the third layer, we propose the use of a pre-trained model based on different ML approaches which may optimize memory, computational usage, and system performance. The whole multi-algorithm architecture constitutes a pre-trained system that can perform segmentation for any correlation matrix, regardless of the nature of the correlation function and its size.

Extensive experiments have demonstrated the superior performance of the proposed approach in synthetic databases. We have shown that a Ridge Regression algorithm obtains an excellent trade-off between performance, memory size, and speed, improving the processing speed and in-memory size by 4 orders of magnitude over the SVR, and 1 order of magnitude over the MLP, with a performance loss of less than 3%. In addition, a real problem has been proposed based on the correlation that a Natural Language Model (BERT) provides between sentences for text segmentation, which is one of the fields where it is useful to segment correlation matrices. Unlike most state-of-the-art algorithms, the proposed architecture can work with different scales and input sizes with high performance and speed. Using this real problem, we have also presented a performance comparison of segmentation obtained with other state-of-the-art models, improving 5.43% the *MSE* over the second-best tested algorithm and 36.35% the *WD* performance. The proposed approach has been completely programmed in Python, and it is freely available to be used and adapted for different correlation matrix segmentation problems and can be compared with alternative new approaches.

| Type | Model | | Wikipedia (%) | | Synthetic (%) | | Size (kB) | Comp. time (s./7.5 kS) | # par. |
|---|---|---|---|---|---|---|---|---|---|
| | | | 1↓ WD | 1↓ MSE | 1↓ WD | 1↓ MSE | | | |
| CD | Louvain | [18] | 36.16 | 75.60 | 28.18 | 45.80 | 0 | 8.27 | 1 |
| CD | MM | [27] | 37.28 | 77.23 | 76.12 | 83.59 | 0 | 24.50 | 1 |
| US | SegCorr | [21] | 60.58 | 91.48 | 76.97 | 86.28 | 0 | 18.30 | **1** |
| US | HC | [26] | 55.48 | 90.78 | 48.56 | 67.87 | 0 | 0.61 | 2 |
| US | SC | [23] | 60.94 | 90.22 | 80.21 | 88.31 | 0 | 15.69 | 2 |
| US | PBMM | [8] | 76.87 | 94.56 | 64.89 | 80.45 | 0 | **0.21** | 3 |
| DC | MLP-DC | [64] | 41.19 | 79.12 | 64.56 | 76.57 | **35.05** | 91.65 | 297 |
| DC | CNN-DC | [64] | 58.92 | 91.24 | 58.85 | 76.18 | **35.05** | 204.65 | 297 |
| | CoSeNet | | **82.60** | **96.83** | **87.19** | **92.47** | 131.7 | 3.13 | 5 |

Table 5: Baseline comparison between Community Detection (CD), unsupervised (US), Deep Clustering (DC) and our proposal in terms of performance (1↓ WD) for a synthetic database with noise variance of 0.2 ($\omega^2 = 0.2$) and the Wikipedia database, memory size, computational speed (Comp. time) and number of parameters (# par.) to optimize. The best solution between state-of-the-art algorithms and the proposed CoSeNet architecture is shown in bold.

The application of this model is to divide a correlation matrix into its possible segments. However, a large number of ordered clustering problems can be transformed into idenfying segments in a correlation matrix, making it a potential problem-solving tool. For example, apart from the NLP problem, SME Portfolio Segmentation works with correlation matrices, and this model is able to identify the segments of such matrices. For cis-regulatory modules, there is a function that correlates DNA features. The values can be arranged into a correlation matrix, and gene regions can be identified by CoSeNet. In summary, whenever there is a clustering problem with ordered entities, this model is fast and efficient in segmenting groups of entities.

*5.1. Model limitations and further work*

Regarding the main limitations of the proposed model, note that its performance is highly dependent on the correlation function generated by the input matrices, which is highly dependent on the problem itself and its applicability. Due to the proposed model architecture, the noise characteristics that affect the segmentation are only variance and the non-linearity of the correlation method. That is why we have included in the architecture a specific layer capable of correcting the non-linearity of the matrices, as long as the user has a dataset that can allow fine-tuning the model parameters. Unfortunately, the noise variance itself is insurmountable, and it is the most limiting aspect to obtaining better results in general.

As possible improvements to the proposed model, filter banks that correct specific types of noise in the input correlation matrices could be introduced as part of an initial pre-processing step. This filtering would greatly facilitate the task of segment prediction. New scaling functions could also be proposed to improve the non-linearity correction. Regarding the ML model implemented, it is hard to find a model with a better trade-o! than Ridge regression because it is a linear regressor that runs with simple matrix multiplication, and has a perfect hit rate for segmenting correlation matrices without noise. Any further analysis with Deep Learning approaches, for example, will be carried out at the expense of hardly extending the computational cost of the algorithm.

**Acknowledgements**

This work has been supported by *Universidad de Alcalá - ISDEFE* Chair of Research in ICT and Digital Progress. This research has also been partially supported by the project PID2020-115454GB-C21 of the Spanish Ministry of Science and Innovation (MICINN).

**Code**

The proposed multi-algorithm architecture code (CoSeNet and data for the experiments) used in this paper are available at: `https://github.com/iTzAlver/CoSeNet`.